\documentclass[10pt,twocolumn,letterpaper]{article}

\usepackage{wacv}
\usepackage{times}
\usepackage{epsfig}
\usepackage{graphicx}
\usepackage{amsmath}
\usepackage{amssymb}

\usepackage{enumitem}

\wacvfinalcopy 

\ifwacvfinal
\def\assignedStartPage{1} 
\fi

\ifwacvfinal
\usepackage[breaklinks=true,bookmarks=false]{hyperref}
\else
\usepackage[pagebackref=true,breaklinks=true,colorlinks,bookmarks=false]{hyperref}
\fi

\ifwacvfinal
\else
\fi

\begin{document}
\title{FIR-based Future Trajectory Prediction in Nighttime Autonomous Driving}

\author{Alireza Rahimpour$^1$, Navid Fallahinia\thanks{Equal contribution} $ ^ 2$, Devesh Upadhyay$^3$, Justin Miller$^3$\\
 \\
$^1$Ford Motor Company, Greenfield Labs, Palo Alto, CA, USA\\
$^2$University of Utah,
Salt Lake City, UT, USA\\
$^3$Ford Motor Company, Dearborn, MI, USA}

\maketitle

\begin{abstract}
   The performance of the current collision avoidance systems in Autonomous Vehicles (AV) and Advanced Driver Assistance Systems (ADAS) can be drastically affected by low light and adverse weather conditions. Collisions with large animals such as deer in low light cause significant cost and damage every year. In this paper, we propose the first AI-based method for future trajectory prediction of large animals and mitigating the risk of collision with them in low light. 
   In order to minimize false collision warnings, in our multi-step framework, first, the large animal is accurately detected and a preliminary risk level is predicted for it and low-risk animals are discarded. In the next stage, a multi-stream CONV-LSTM-based encoder-decoder framework is designed to predict the future trajectory of the potentially high-risk animals. The proposed model uses camera motion prediction as well as the local and global context of the scene to generate accurate predictions. Furthermore, this paper introduces a new dataset of FIR videos for large animal detection and risk estimation in real nighttime driving scenarios. Our experiments show promising results of the proposed framework in adverse conditions. Our code is available online\footnote{\url{https://github.com/FordCVResearch/FIR-Trajectory-Prediction}}. 
\end{abstract}

\section{Introduction}

An estimated 1 to 2 million crashes between motor vehicles and large animals, such as deer,  occur every year in the U.S., causing approximately $200$ human deaths, 26,000 injuries, and at least $8$ billion in property damage and other costs \cite{williams2005characteristics}. National Safety Council statistics have shown that accident rates are three times higher at nighttime compared to daytime \cite{6856446}.
Darkness and adverse weather dramatically reduce the distance at which drivers can see and react to approaching hazards. 
Not only will the driver's perception ability be affected in darkness, but many onboard driver's assist sensing equipment will not be able to detect collisions under low visibility. 

In contrast to these limited existing perception systems, FIR cameras provide images with high contrast between living objects (e.g., animals) and vegetation and similar cover. The amount of thermal radiation emitted by an animal like a deer is much greater than the surrounding cover, making robust detection possible. FIR cameras work in all weather conditions such as fog, snow, and rain. Moreover, they work in all ambient lighting conditions such as night, day, and when bright visible lighting is present (e.g., due to oncoming headlights, sun glare, or tunnels, etc.) Therefore, the FIR camera shows promise to be one of the most reliable sensors for large animal detection and collision avoidance systems in ADAS/AV. 
Furthermore, a collision avoidance system in ADAS/AV should be very robust and must meet the ASIL level safety requirements that mandate very low false alarm rates for the Automatic Emergency Braking (AEB) system. The same is true for a warning system or a trajectory planning system in AV. Therefore, estimating the future trajectory of the animal for collision risk prediction is critical. 

\begin{figure*}[h!]
    \centering
    \includegraphics[width=\linewidth]{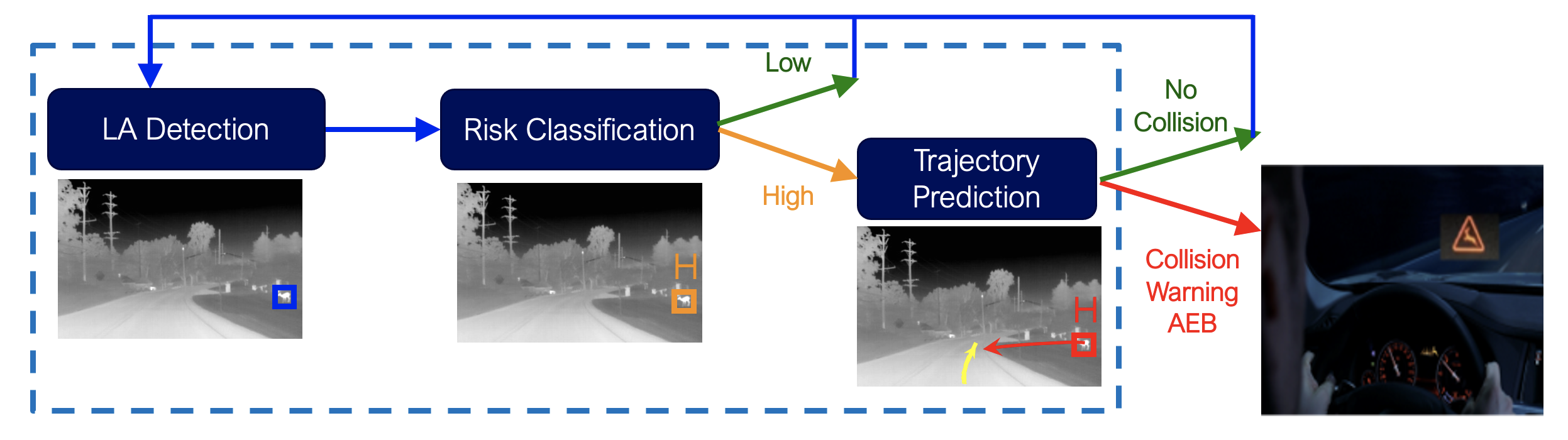}
    \caption{The proposed multi-step framework: Large Animal 
    Detection (LAD)- early risk assessment-trajectory prediction. The warning will be shown in HUD and AEB will be activated only when there is a high risk of collision based on the predicted trajectories.}
    \label{fig:1}
\end{figure*}
To the best of our knowledge, there is no similar FIR-based large animal future trajectory prediction for collision risk estimation platform. The closest works to ours in this field are \cite{WinNT, 1252663,6856446, rahimpour2022enhanced}. 
\cite{WinNT} (i.e., the most recent work) is the Large Animal Detection (LAD) system by Volvo which is limited and does not work in harsh lighting and weather condition. Moreover, it cannot detect large animals when they are partially obscured or have limited contrast to their immediate background. The camera should also be able to see the animal straight from the side and assumes that the animal has a normal pattern of movement \cite{WinNT}. The proposed framework in this paper addresses all of these limitations. 
In \cite{6856446, 1252663}, an animal detection system has been proposed based on using rule-based methods and traditional features. Each component of the model is designed based on physical constraints and strong assumptions about the animal and environmental properties which can be easily violated in the real world. 
Unlike, \cite{WinNT, 6856446, 1252663} our model can automatically learn useful contextual features effectively which leads to a more robust detection and prediction in different temperatures and environments. 
Moreover, \cite{6856446, 1252663, WinNT,rahimpour2022enhanced} are only based on animal detection and do not perform any large animal future trajectory estimation. 

Recently there have been some perception-based approaches for predicting the future intention of pedestrians \cite{ahmed2019pedestrian, fang2017board, kooij2014context, rasouli2017they, ridel2018literature, rasouli2019pie, malla2020titan, gupta2018social, rahimpour2021system}. One approach is based on dynamical motion modeling (DMM) \cite{keller2013will}.
The head and body orientation have been also used for pedestrians' intention estimation \cite{schulz2015pedestrian, huang2013head}. 
However, these works are limited only to intent prediction for pedestrians, which is simpler than large animals since pedestrians have fewer sudden movements compared to deer and they typically follow established paths and follow traffic signals and rules \cite{6856446}.
Moreover, animals in traffic scenarios are often occluded by vegetation. Especially our dataset showed that deer's legs are occluded by grass in most cases. 
Further, unlike pedestrians, the animals’ appearance varies substantially depending on the point of view. 
Moreover, none of the mentioned methods would work in nighttime and adverse weather, nor do they consider the estimation of the onboard camera's future movement, low image resolution (i.e., FIR images), fast movements, and heavy occlusion of the deer in their intent prediction. 
To summarize the main contributions of this paper are as follows:
\begin{itemize}

\item Designing a novel memory-based large animal trajectory prediction for nighttime and adverse weather. The proposed CONV-LSTM-based framework is able to predict complex patterns and works for both stationary and fast-moving deer in various pose and occlusion scenarios. 

\item  Proposing a multi-step framework that minimizes the possibility of false collision warnings thanks to multiple filtering of the input by accurate and robust detection, early risk classification, and incorporating the future motions of the camera on board the ego-vehicle. 

\item Collecting a new dataset for large animal detection and prediction with accurate human annotations and metadata. One of the biggest challenges in the detection and prediction of large animals is the lack of data for training. 

\item Systematic quantitative and qualitative evaluation of the proposed method against baselines. 
\end{itemize}
%

\section{Method} 
\subsubsection*{Multi-Step Framework}
Our proposed model is composed of three stages as shown in Figure \ref{fig:1}.
\begin{figure*}
    \centering
    \includegraphics[width=\linewidth]{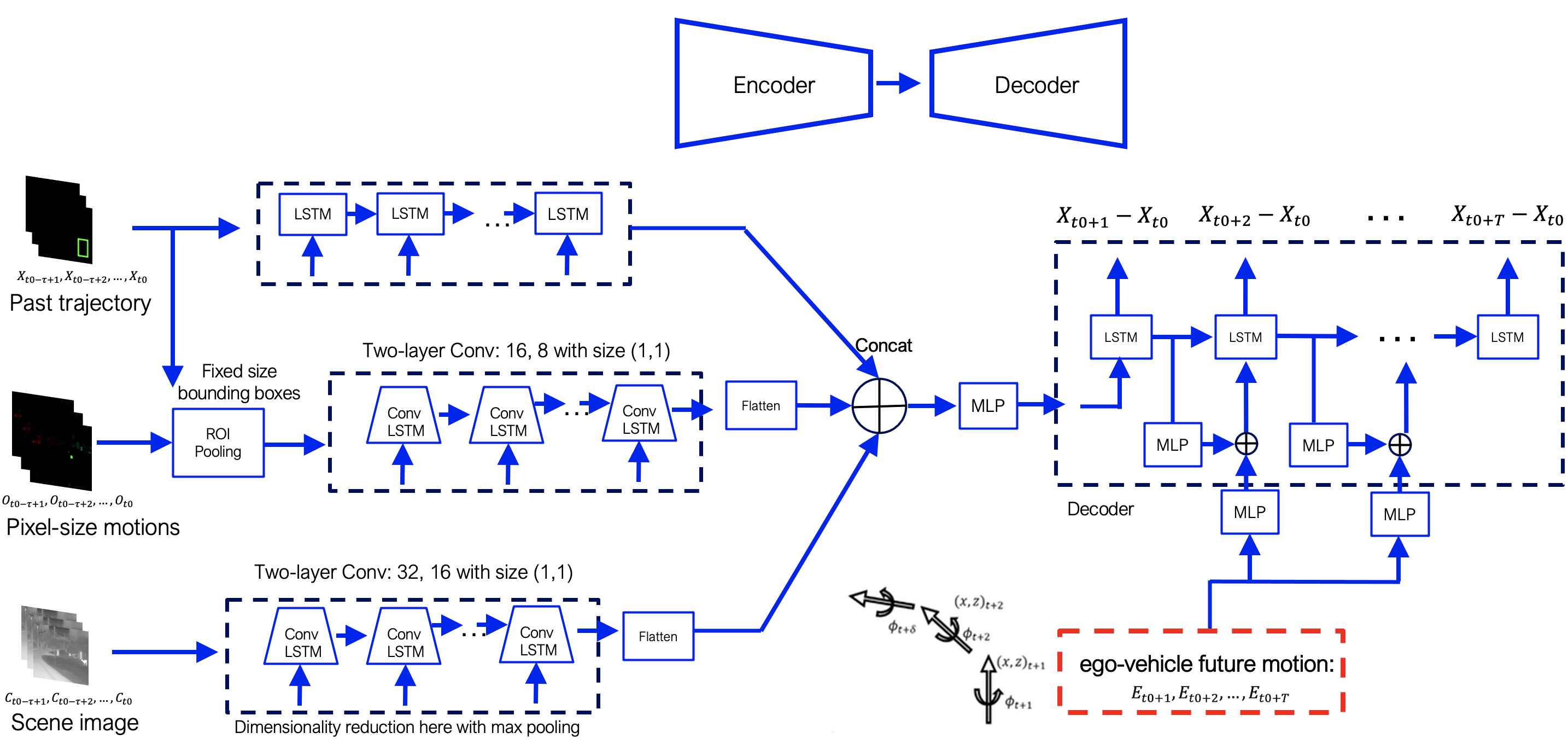}
    \caption{The details of the proposed future trajectory prediction model architecture. In our proposed model the future motion of the FIR camera on board the moving ego vehicle is also estimated (red box) and is incorporated into the decoder.}
    \label{fig:2}
\end{figure*}
In the first stage, an FIR-based large animal detection model is trained to detect all visible animals (deer in this study) in the scene at both close and far distances. For training the detection model, FIR image sequences are collected and annotated with a bounding box around each deer and a label (deer/not deer).

The second stage is the preliminary risk classification based on risk level labels \cite{8814210}. Each large animal in the images is annotated with risk level (i.e., High/Low risk) based on its pose, trajectory, and distance to the road/ego vehicle. The model predicts the  preliminary risk level for each detected animal in each frame and only selects the animals with high risk for processing in the next stage. 
Filtering the images in this way increases the accuracy of the model and reduces the computational cost since we only process the potentially dangerous cases and ignore the cases where the animal has low collision risk (e.g., is very far, etc.). Since FIR cameras are capable of detecting large animals at far distances, this step is necessary to reduce the cognitive load of the system by only focusing on animals that are close to the roadside.
The third stage is the prediction of the animal's future trajectory/location to determine the possibility of any future collision with the vehicle. We developed a novel Conv-LSTM multi-stream encoder-decoder model for predicting the animal's future location based on a sequence of past FIR video observations. Our proposed model is memory-based and preserves both temporal and spatial information and it only requires two seconds of observation time to make predictions about the animal’s trajectory/location in the future. Thus, it can make predictions within two seconds of detecting any new animal in the camera's field of view. 
In order to make the proposed collision avoidance system more robust against sudden changes in animal  crossing behavior, the entire process of detection, risk classification, and trajectory prediction is continuously repeated in a tracking loop. Our proposed framework ensures minimum false collision warning by filtering out false positives in each stage of the model. 

\subsubsection*{Multi-Stream Trajectory Prediction Model}
The trajectory prediction model is designed to benefit from multiple input streams in order to maximize its observation domain. The model architecture is shown in Figure \ref{fig:2}. The three input streams to the model are the past trajectory ($X$), pixel-level motions ($O$), and scene context ($C$) in ($t_0 - \tau + 1, ..., t_0$) and the goal is to predict the bounding box location around the animal after $t_0$. 

The first input is the past trajectory in the form of the 2D location of the detected bounding boxes from the previous two-second observation ($60$ frames), which provides information about the object’s trajectory history in the past. This data also provides valuable information about the animal’s average velocity and heading in the past. 
This vector goes through the LSTM-based branch of the encoder.

The second input is pixel-level motions. 
We use dense optical flow for the pixels inside each detected bounding box which provides precise information about the motion of each pixel with respect to the background. This data provides the prediction model with information about the pixel-level movements of the animal body. This input first goes through the ROI pooling layer, and then CONV-LSTM layers are used to encode them.

The third input to the model is the global context which provides valuable information about the scene, such as proximity to the road elements, road curves, etc. Due to some unpredictable motions and sudden movements by the animal, it is not possible to only rely on the animal’s motion history to make predictions about its future trajectory/location. Therefore, it is essential to search the scene for elements that might prevent the animal from crossing the road. This is inspired by the human brain’s prediction mechanism, where both high-level and low-level changes in the scene are used to make a prediction. Our experiments showed that incorporating this global context into the model significantly helps in predicting the sudden movements of the deer. 
CONV-LSTM layers are used to encode the whole scene which is used as a global context. Each of the encoders will generate its encoded context vector associated with its domain (left side of Figure \ref{fig:2}). 

The encoded feature vectors from these three branches are flattened and concatenated. The features can be concatenated using either a Transformer \cite{vaswani2017attention} or Multi-Layer Perceptron (MLP).  We used an adaptive transformer-based fusion but it did not improve the results. Thus, for the experimental results, we only show the performance of the MLP-based model.
The encoded features from the observed sequence then go into the LSTM-based decoder and the output of the decoder is the future trajectory of the large animal after $t_0$ (right-hand side of Figure \ref{fig:2}). It is important to note that since the camera is on-board, the ego-vehicle state (location, heading) must be considered to compensate for the relative motion between the object and the vehicle (i.e., camera). We use visual odometry (ORB-SLAM) \cite{mur2015orb} to estimate the vehicle's future motion from first-person view images. The main decoder will use both the context feature from the past observation and the predicted vehicle’s future motion as inputs to make predictions about the future trajectory of the large animal. Note that the vehicle state i.e., $E_{t_0 + 1}, ...$ is predicted for the future time (after $t_0$) and hence it is incorporated into the decoder. More details about the overall architecture of the model can be seen in Figure \ref{fig:2}.

\section{Experiments}
\textbf{Dataset:}
There is no publicly available data for FIR-based large animal detection and trajectory prediction. Hence, we collected and annotated $26,127$ FIR images in rural areas of Michigan in real nighttime driving scenarios. The collected dataset includes both moving and stationary animals. Also, there are many edge cases and occluded deer in the dataset which are very helpful for evaluating collision estimation algorithms. 
The FIR camera is mounted in the grille of a Ford F150.  
A FLIR Boson camera with $50$ degrees horizontal field of view is used for data collection. The images have VGA ($640 \times 480)$ quality and $30$ frames are collected per second. There are $77,291$ bounding boxes in our collected data. Each deer in the dataset is annotated with a bounding box and a risk level label (high/low). Figure \ref{fig:3} shows some samples of the collected dataset and their annotations in green.

\textbf{Model Implementation}
For the first stage of our framework, Faster RCNN \cite{ren2015faster} with ResNet101 backbone is used for large animal detection. A classification head is added to FasterRCNN for early risk estimation (low/high). For the trajectory prediction, as shown in Figure \ref{fig:2}, we employed two stacks of $60$ Vanilla LSTM blocks for each of the encoders (the top branch) and the decoder. We used 2 layers of Conv-LSTM for the rest of the encoder. The entire model was trained end-to-end using a stochastic gradient descent optimizer.
We use a mean squared error loss function, which is calculated by picking the top $5$ trajectories closest to the ground truth from a sample of $10$ future trajectories. 
Also, for faster convergence, we used normalized values of the bounding box centers and aspect ratios. The proposed model was implemented in Pytorch and trained on two NVIDIA TITAN RTX GPUs with $300$ epochs. 
For more details about the architecture and hyperparameters please refer to the GitHub page of this project. 

\textbf{Evaluation Metrics}. 
Previous related works \cite{6856446, WinNT} have not reported their evaluation metrics and results. Regardless, we use Average Precision (AP) for our evaluation of large animal detection and risk estimation.  
For trajectory prediction, we follow works in pedestrian trajectory prediction. 
Following \cite{alahi2016social}, we considered two error metrics to evaluate the performance of the model: 
\begin{enumerate}[noitemsep,topsep=0pt,parsep=0pt,partopsep=0pt]
\item Average Displacement Error (ADE): The Average $L2$ distance between ground-truth and predicted bounding-box center for the duration of the prediction.
\item Final Displacement Error (FDE): The L2 distance between the ground truth and the predicted bounding-box center at the final frame
\end{enumerate}
For evaluation, we train on 4 sets of data and test on 1 as a leave-one-out strategy. Each training step is observed for 2 seconds and prediction errors are measured for 1 second. The reported errors are normalized. 
\subsection{Results}

\subsubsection{Quantitative analysis}
The test set for large animal detection includes $6000$ images. Our importance-guided training scheme for large animal detection and risk classification showed very promising results leading to AP of $95\%$. The results showed a very robust performance of our model even for very far and occluded animals in different poses (Figure \ref{fig:3}). 

\textbf{Baseline and ablation study}
We use the vanilla social LSTM model \cite{alahi2016social} (with no social pooling) which is similar to the top branch of our model (when only using past trajectory as input and LSTM-based encoder-decoder with no vehicle/camera motion prediction) as a baseline. To prove the effectiveness of each part of our model we also performed the ablation study for the following cases: LCV: Location-Context-Vehicle/camera motion (no pixel motion), LMV: Location-Motion-Vehicle/camera motion (no context), LMC:Location-Motion-Context
(no vehicle/camera motion prediction), and \textbf{LMCV}: the full model. 

Table \ref{table:1} shows the quantitative analysis results in terms of the average FDE and ADE. From these metrics, it can be observed that the proposed model (LMCV) outperforms all the variations in predicting the future trajectory of the deer (the center of the bounding boxes). 
Furthermore, our results confirm the effectiveness of each module in our proposed framework. 
Results show that only using LSTM without taking advantage of pixel-level motions and context would lead to the worst results. The performance of the LCV model in Table \ref{table:1} shows the importance of directly using the dense optical flow for accurate estimations of sudden deer movements.   
We can also observe a significant error increase in the LMC model where the effect of the future motion of the camera is removed. This shows the importance of calculating the relative movement of the deer with respect to the ego-vehicle and camera. The LMV model confirms the importance of the global context of the scene (driving environment, trees, etc.) in predicting the deer trajectory. 
\begin{table}[h!]
\centering
    \caption{Baseline and Ablation study: comparison of the performance of the full model (LMCV) with baseline and other variations by removing one component in each experiment. The lower error is better.}
    \label{table:1}
\begin{tabular}{|c|c|c|c|c|c|}
\hline
Metric & Baseline \cite{alahi2016social} & LCV & LMV & LMC& \textbf{LMCV} \\ \hline
ADE           & 0.42& 0.30 & 0.21 & 0.27 & \textbf{0.13}\\ \hline
FDE          &0.69 & 0.42 & 0.29 & 0.32 & \textbf{0.18} \\ \hline
               
\end{tabular}
\end{table}
Additionally, we test the performance of the LMCV model for different prediction duration. Table \ref{table:2} shows that the prediction performance degrades when the prediction time is larger than the observation time (i.e., 2s). 
\begin{table}[h!]
\centering
    \caption{The ADE, FDE across different test duration.}
    \label{table:2}
\begin{tabular}{|c|c|c|c|}
\hline
Average Error & 1 second & 2 seconds & 4 seconds \\ \hline
ADE           &0.13 & 0.31 & 0.41 \\ \hline
FDE           &0.18 & 0.36 & 0.52 \\ \hline
\end{tabular}
\end{table}

\subsubsection{Qualitative results}
\begin{figure}[h!]
    \centering
    \includegraphics[width=1\linewidth]{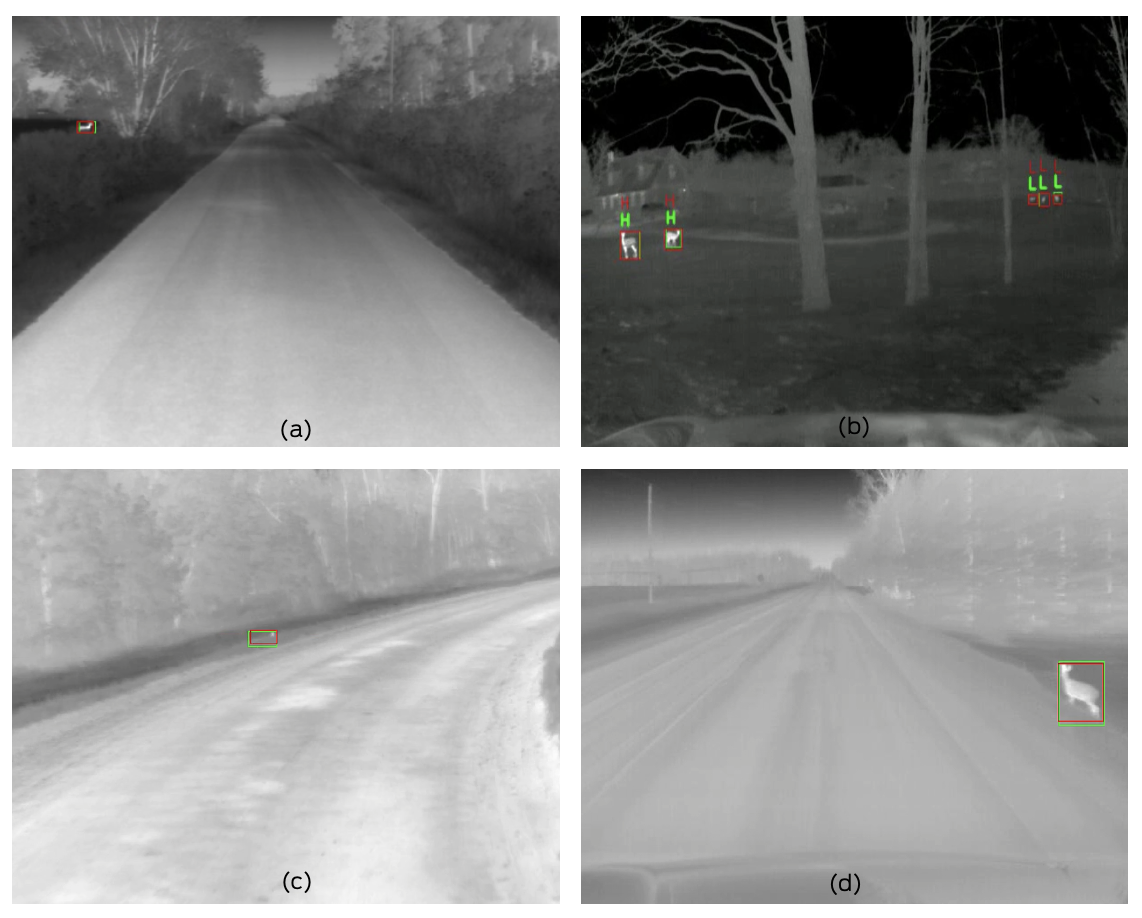}
    \caption{Samples of the collected dataset and detection results of our model in heavy occlusion (a, c), different animal poses, and far distances (a, b, d). Green is the ground truth and red is the model estimation.}
    \label{fig:3}
\end{figure}

\begin{figure}[h!]
    \centering
    \includegraphics[width=1\linewidth]{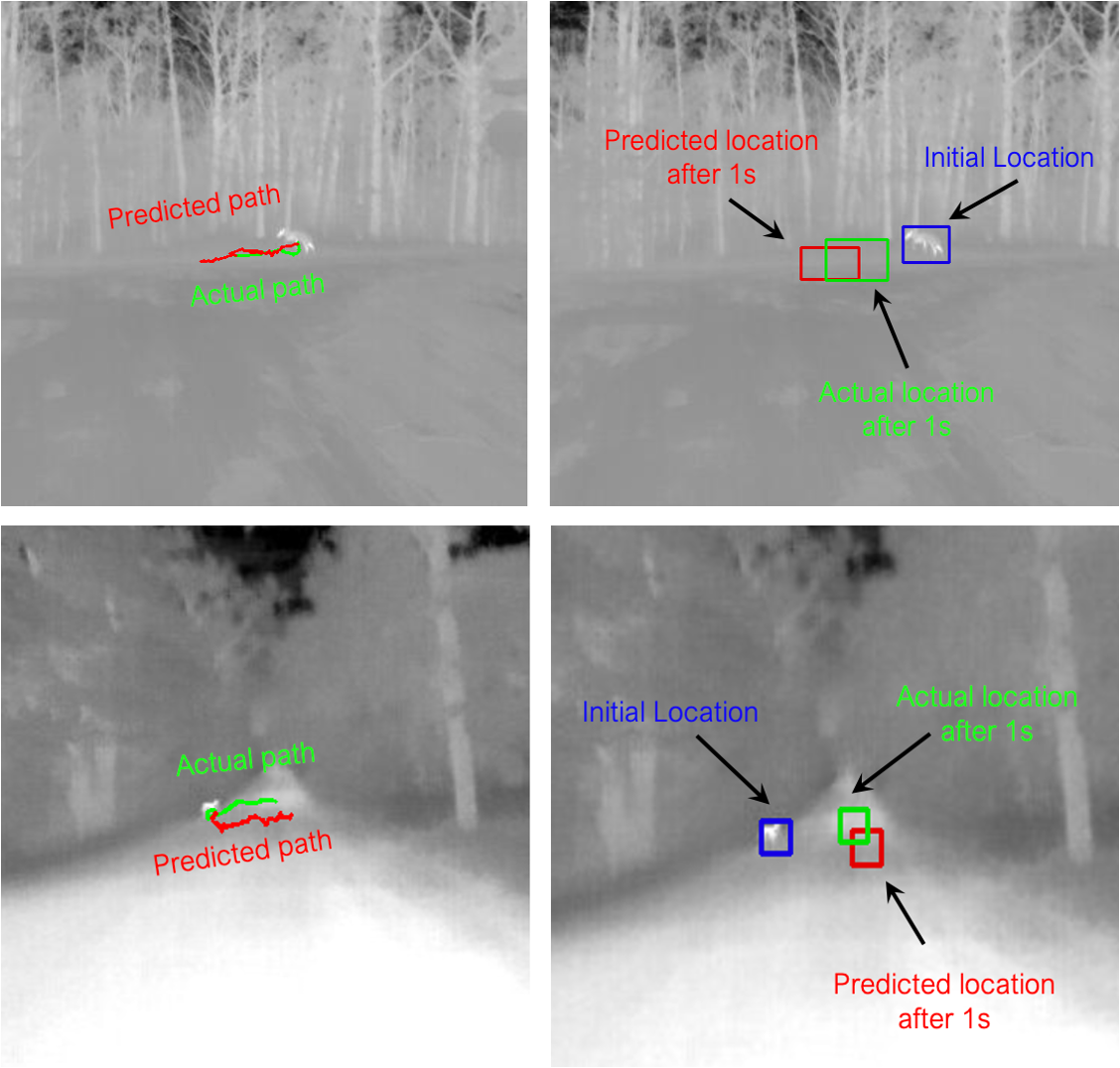}
    \caption{Results of the proposed large animal trajectory prediction for a fast and sudden movement of the animal in 1 second.}
    \label{fig:4}
\end{figure}

In addition to the quantitative results, some qualitative results of the detection, risk estimation, and future trajectory prediction are shown in Figure \ref{fig:3} and Figure \ref{fig:4}.
The trained detection and risk estimation model shows great performance in various scenarios. For example, Figure \ref{fig:3} (a-c) shows the power of the model to deal with occlusion and different animal poses. In Figure \ref{fig:3} (c) we can see that our model can assign a low-risk label to deer in the far distance and ignore them in the next stage of the model and save computational cost. Moreover, the robustness of the model with respect to the camera mounting location on the car can be observed (the camera location is different in a,c vs. b,d). 
The qualitative results in Figure \ref{fig:4} give some perspective to the way in which prediction errors manifest. In this scenario, the deer jumps in 1 second and our model can successfully predict the future trajectory even though the deer is far away and partially occluded (legs are not visible). We see that while the predicted future bounding box location is not perfect, the overall direction of travel and approximate future large animal location are sufficiently accurate to avoid a collision.
Despite the fact that predicting the complex movements of the deer is challenging in the FIR domain with low resolution and lack of texture, our framework demonstrated promising performance.

\section{Conclusion and Discussion} 
We proposed the first FIR-based framework for predicting the future trajectory of large animals (i.e., deer) in nighttime driving scenarios.  
We showed that our framework is able to successfully compensate for the relative motion of the camera and deer during the prediction period thanks to the SLAM-based motion prediction module. 
Having the future trajectory of the large animal and the speed and trajectory of the ego vehicle, the collision risk, and Time To Collision (TTC) are estimated for each animal in the scene and if the collision risk is high, a warning to the driver or the AEB is initiated.
Our work will help address the important challenges with nighttime driving faced by vision-based ADAS and AV systems. No autonomous vehicle can reliably work without taking into account the trajectory of road users in all adverse conditions. 

This work initiates the research on FIR-based large animal trajectory prediction by providing a new dataset and baseline experiments. 
While currently only focused on FIR images, the proposed model can be fused with other sensory inputs such as RGB, radar, or Lidar to increase the accuracy and robustness. For instance, knowledge distillation can be used in our framework for transferring knowledge from a multi-modal teacher model (e.g., FIR + RGB) to a single-modal (e.g., RGB) student model and increasing the robustness of trajectory prediction at nighttime without adding any cost during inference.   
Our current research includes using Graph Neural Networks (GNN) to model the interaction between road users (e.g., deer, etc.) and ego vehicle. Furthermore, we are investigating end-to-end detection and prediction using vision transformers (ViT) and attention mechanism \cite{rahimpour2019attention}, and uncertainty-aware path prediction and planning. Moreover, we believe that common evaluation metrics for detection and prediction (such as mAP, ADE, FDE, etc.) are not representative of the actual performance of the models, and more task-specific metrics need to be defined. 

\textbf{Acknowledgement }
We would like to thank Jonathan Diedrich and Mark Gehrke at Ford Motor Company for their efforts in collecting the FIR-based large animal detection dataset. 


{
\small
\bibliographystyle{ieee}
\bibliography{ref3}
}

\newpage
\appendix


\end{document}